\documentclass[letterpaper, twocolumn, 10pt, conference]{ieeeconf}  

\IEEEoverridecommandlockouts                             

\usepackage{ifpdf}
\usepackage{cite}
\usepackage{graphicx}
\usepackage{amsmath, amssymb}
\usepackage[linesnumbered, ruled]{algorithm2e}
\usepackage{array}
\usepackage{mdwmath}
\usepackage{mdwtab}
\usepackage{eqparbox}
\usepackage{subcaption}
\usepackage{stfloats}
\usepackage{url}
\usepackage{multirow}
\usepackage{multicol}
\usepackage{mathtools}
\usepackage{color}
\usepackage{flushend}

\newcommand{\by}{{\mkern-2mu\times\mkern-2mu}}
\newcommand{\norm}[1]{{\left\lVert#1\right\rVert}}

\newcommand{\SO}{\rm{SO}}
\newcommand{\SE}{\rm{SE}}

\newcommand{\se}{\mathfrak{se}}
\newcommand{\skls}{\SE(3)^n}

\title{\LARGE \bf
Signal Alignment for Humanoid Skeletons via the Globally Optimal Reparameterization Algorithm
}

\author{Thomas W. Mitchel*, Sipu Ruan*, Gregory S. Chirikjian
\thanks{*Equally contributing authors} 
\thanks{Thomas W. Mitchel, Sipu Ruan and Gregory S. Chirikjian are with the Laboratory of Computational Sensing and Robotics, Johns Hopkins University, Baltimore, Maryland 21218 {\tt\small \{tmitchel, ruansp, gchirik1\}@jhu.edu}}}

\begin{document}
\maketitle
\thispagestyle{empty}
\pagestyle{empty}
\begin{abstract}
The general ability to analyze and classify the 3D kinematics of the human form is an essential step in the development of socially adept humanoid robots. A variety of different types of signals can be used by machines to represent and characterize actions such as RGB videos, infrared maps, and optical flow. In particular, skeleton sequences provide a natural 3D kinematic description of human motions and can be acquired in real time using RGB+D cameras. Moreover, skeleton sequences are generalizable to characterize the motions of both humans and humanoid robots. The \textit{Globally Optimal Reparameterization Algorithm} ({\bf GORA}) is a novel, recently proposed algorithm for signal alignment in which signals are reparameterized to a globally optimal \textit{universal standard timescale} ({\bf UST}).  Here, we introduce a variant of GORA for humanoid action recognition with skeleton sequences, which we call \textbf{GORA-S}. We briefly review the algorithm's mathematical foundations and contextualize them in the problem of action recognition with skeleton sequences. Subsequently, we introduce GORA-S and discuss parameters and numerical techniques for its effective implementation. We then compare its performance with that of the {\it DTW} and {\it FastDTW} algorithms, in terms of computational efficiency and accuracy in matching skeletons. Our results show that GORA-S attains a complexity that is significantly less than that of any tested DTW method. In addition, it displays a favorable balance between speed and accuracy that remains invariant under changes in skeleton sampling frequency, lending it a degree of versatility that could make it well-suited for a variety of action recognition tasks. 

\end{abstract}

\section{INTRODUCTION} \label{intro}

As the name suggests, humanoid robots inherently resemble human bodies, and typically consist of a head, torso, two arms, and two legs. Moreover, many of these robots are designed to emulate human behaviors, such as walking or dancing, and to communicate with us both verbally and non-verbally, through conversation or universal gestures such a wave or a cheer \cite{kaneko2008humanoid}. Humanoid robots have the potential to perform a variety of challenging tasks normally reserved for their biological counterparts, such as providing office or administrative support in the role of a receptionist or caring for the elderly; as such, they have become a popular area of research in robotics. 

\begin{figure}
\centering
\includegraphics[width= \columnwidth]{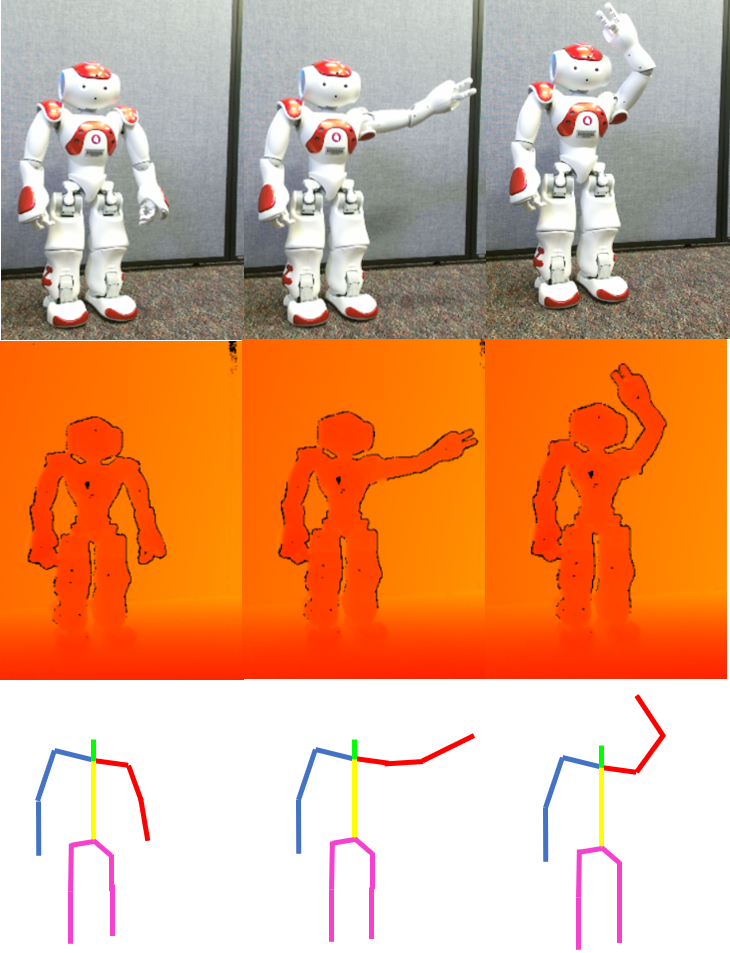}
\caption{A wave by a NAO robot recorded as three different data types. $1^{st}$ row: RGB video; $2^{nd}$ row: infrared depth map; $3^{rd}$ row: skeleton sequence.  A robust action recognition should have the ability to synthesize and compare signals of different data types to correctly categorize a given action.}
\label{fig:demo_nao_multimodal}
\end{figure}

Socially adept humanoid robots require a detailed knowledge of human actions in the context of daily life, and a first step toward this end is the recognition and classification of specific actions using signals or sequences. With the recent emergence of new machine learning and computer vision techniques, new robotic action recognition methods have been developed for a variety of different types of signals, such as sequences of RGB images \cite{ma2018region}, infrared maps \cite{akula2018deep} and 3D skeletons \cite{liu2018skeleton}. The foundation of action recognition lies in the problem of signal alignment, in the sense that prior to categorizing sets of sequences, one should be able to temporally reparameterize the sequences in such a way that enables standardized comparisons. In particular, a robust action recognition algorithm should also be adaptable for use with different modalities of data. For example, such an algorithm should be able to accurately characterize an action, such as a wave from a NAO humanoid robot, by synthesizing and comparing multiple sequences of different data types, as shown in Fig. \ref{fig:demo_nao_multimodal}.

The \textit{Globally Optimal Reparameterization Algorithm} ({\bf GORA}) is a novel, recently proposed algorithm for signal alignment in which signals are reparameterized to a \textit{universal standard timescale} ({\bf UST}) using principles of variational calculus \cite{mitchelruan2018,chirikjian2017signal}. GORA has been initially applied to compare RGB video sequences \cite{mitchelruan2018}, and has shown potential in providing a highly-effective framework for signal comparisons as an alternative to the well known DTW algorithm \cite{sakoe1978dynamic} and its variants\cite{keogh2001derivative,salvador2007toward,keogh2005exact}. In this paper, we introduce a variant of GORA for humanoid action recognition with skeleton sequences, which we call \textbf{GORA-S}. Skeleton sequences can be viewed as trajectories in the Lie group $\skls \cong \SE(3) \times \SE(3) \times ... \times \SE(3)$ \cite{vemulapalli2014human} and provide a natural 3D kinematic description of the motions, gestures, and actions of both natural and artificial humanoids. As such, action recognition with skeleton sequences is immediately relevant to current research in humanoid robotics as the general ability to analyze and classify the 3D kinematics of the human form will likely play an integral role in the development of socially adept machines. 

In this context, the main contribution of GORA-S is \textit{a framework for signal alignment and action recognition using skeleton sequences that attains a significantly lower computational complexity than DTW methods, in addition to displaying a favorable balance between speed and accuracy that remains invariant under changes in the temporal sampling frequency of the input skeletons}. The remainder of the paper is organized as follows:
First, we provide a brief review of the mathematical foundations of the GORA framework and introduce GORA-S in the context of action recognition with skeleton sequences. Next, we review the algorithm and discuss the chosen parameters and numerical techniques used in its application. We then provide a verification of the algorithm by comparing its performance relative to the DTW and FastDTW algorithms \cite{salvador2007toward}, in terms of both computational efficiency and accuracy in matching skeleton sequences from the NTU RGB+D Action Recognition Dataset \cite{shahroudy2016ntu}, and discuss the results.  We conclude with a remark on the computational significance of the differences between GORA-S and DTW methods, in addition to the authors' plans for the continued development of GORA-S.

\section{PROBLEM STATEMENT AND MATHEMATICAL FOUNDATIONS}
The GORA algorithm is based on the notion that any temporal misalignment between two arbitrary signals can be compensated for by reparameterizing each to a universal standard timescale (\textbf{UST}).  Such temporal parameterizations are differentiable strictly monotonically increasing functions on the unit interval and the set of all such functions forms the \textit{Temporal Reparameterization Group} (\textbf{TRG}), denoted as $\cal{T}$, under the operation of composition of functions. \cite{mitchelruan2018, chirikjian2017signal}.

The foundation of this approach lies in the fact that for a cost function of the form 
\begin{equation}
f(\tau,\dot{\tau}) = \dot{\tau}^2 \mathfrak{g}(\tau), \label{cost}
\end{equation}
where $\tau \in \cal{T}$ and $\mathfrak{g}: \mathbb{R} \, \rightarrow \, \mathbb{R}_{>0}$ is differentiable,
the solution to the Euler-Lagrange equation
\begin{equation}
\frac{\partial f}{\partial {\tau}} -
\frac{d}{dt} \left(\frac{\partial f}{\partial \dot{\tau}} \right) = 0,
\label{eulerlag}
\end{equation}
denoted as $\tau^* \in \cal{T}$, globally minimizes the functional 
\begin{equation}
J = \int_{0}^{1} f\left(\tau,\dot{\tau},t\right)\,dt,
\label{eq:cost_functional}
\end{equation}
where $\dot{\tau} = d\tau/dt$. In addition, the globally optimal solution, $\tau^*$, is unique and can be recovered by inverting the function \begin{equation} F(\tau^*) \doteq \frac{1}{c} \int_{0}^{x^*} \mathfrak{g}^{\frac{1}{2}}(\sigma)\,d\sigma = t, \label{invF} \end{equation}
where $t$ is the temporal variable and 
$$ c = \int_{0}^{1} \mathfrak{g}^{\frac{1}{2}}(\sigma)\,d\sigma. $$ 
A full proof and numerical validation of these results can be found in \cite{mitchelruan2018,chirikjian2017signal}.

In general, any kind of temporally evolving signal, $X(t)$, can be thought of as a mapping from the unit interval to the space $S$, i.e. $X:[0, 1] \, \rightarrow \, S$,  on which that particular type of signal evolves. Defining a metric $d$ on $S$, $(S, d)$ becomes a metric space. Here we consider signals of the form  \begin{align} X(t) =  \left(g_{1}(t), \ g_{2}(t), \ \ldots,  \ g_{n}(t) \right) \in \SE(3)^n,  \label{skelly} \end{align} which we call a \textit{skeletons} when $n > 1$. These types of signals are typically acquired using RGB+D cameras, a well-known example being Microsoft's Kinect. Skeletons can provide detailed descriptions of humanoid motions in 3D space and have become an increasingly popular way to characterize these motions in the context of action recognition due to their ability to be acquired in real time \cite{shotton2011,vemulapalli2014human,shahroudy2016ntu}.

The trick in applying the above results to the problem of action recognition lies in how the function $\mathfrak{g}(\tau)$ is defined. In particular, \textit{it should measure the rate of change of a given signal along the temporal axis}. In this context, the solution to~(\ref{eulerlag}), $\tau^* \in \cal{T}$, is such that the reparameterization of $X(t)$ with respect to $\tau^*$, $X(\tau^*(t))$, globally minimizes the rate of change of the signal. Since $\tau^*$ is unique, it serves the role of a parameterization to a UST, in the sense that for any collection of signals, $X_1(t), \ \ldots, \ X_{k}(t)$, their corresponding globally optimal solutions, $\tau^*_1(t), \ \ldots, \ \tau^*_k(t)$, reparameterize each signal to the same timescale which minimizes their rate of change.

In this paper, we introduce a variant of GORA for humanoid action recognition with skeleton sequences, which we call \textbf{GORA-S}. Given an arbitrary skeleton signal $X(t) \in \skls$ and initial temporal variable $t$ as inputs, GORA-S recovers the UST parameterization $\tau^*(t)$ corresponding to $X(t)$ and outputs the UST reparameterization of the skeleton, $X^*(t) = X(\tau^*(t))$. This UST reparameterization of the input skeleton can then be compared element-wise with other UST reparameterized skeletons to determine whether or not they represent the same action, resulting in a linear complexity of $O(T)$, where $T$ is the total number of time instances in the sequence. 

As an illustration of this method, suppose $X_1(t), X_2(t) \in \SE(3)^n$  are skeletons upon which minimal nuisance parameters or motion artifacts are acting and let $d$ be an arbitrary metric on the $\skls$. Subsequently GORA-S can be used to find the UST reparameterizations of $X_1(t)$ and $X_2(t)$, given by $X_1^*(t) = X_1(\tau_1^*(t))$ and $X_2^*(t) = X_2(\tau_2^*(t))$, respectively. Then, we can say that $X_1(t)$ and $X_2(t)$ represent the same humanoid action or gesture if  $$ \int_0^1 d(X_{1}^*(t), X_{2}^*(t)) \ dt \approx 0, $$ despite any initial temporal misalignment.

\section{THE GLOBALLY OPTIMAL REPARAMETERIZATION ALGORITHIM FOR SKELETON SEQUENCES (GORA-S)} \label{algorithm}
This Globally Optimal Reparameterization Algorithm for Skeleton Sequences (GORA-S) is defined in Algorithm \ref{algo1}.

\begin{algorithm}
\SetKwInOut{Input}{Input}
\SetKwInOut{Output}{Output}

\caption{Globally Optimal Reparameterization Algorithm for Skeleton Sequences (GORA-S)}
\label{algo1}
\Input{Input skeleton sequence $X(t)$; Initial temporal variable $t$}
\Output{UST reparameterization of skeleton sequence $X^{*}(t)$}

Calculate $\dot{X}(t) = dX/dt$ \\
Compute $\mathfrak{g}(t; \ \dot{X}(t))$;\\
$c = $ {\it NumericalIntegration}($\mathfrak{g}^{\frac{1}{2}}(\sigma)$, $[0,1]$); \\
$F(\tau^*) = \frac{1}{c} \ {\it NumericalIntegration}(\mathfrak{g}^{\frac{1}{2}}(\sigma)$, $[0,\tau^*]$); \\
$\tau^*(t) = F^{-1}(t)$; \\
$X^{*}(t) = {\it Interpolation}(X(t), \tau^*(t); \ \dot{X}(t))$;
\end{algorithm}
Given $\mathfrak{g}(t)$, numerically calculating $\tau^*(t)$ is relatively straightforward and can be done efficiently by first computing $F(\tau^*)$ as in~(\ref{invF}) then interpolating $t$ as a function of $F(\tau^*)$ at the query points given by $t$. The UST reparameterization of the input signal, $X^*(t)$, is recovered by interpolating the input signal as function of $t$ at the query points given by $\tau^*$. Additionally, since the temporal derivative of the input signal, \begin{align*} \dot{X}(t) = (\dot{g}_1(t), \ \ldots, \dot{g}_n(t)) \end{align*}is necessary for both the computation of $\mathfrak{g}(t)$ and to perform our chosen method of interpolation to recover the UST reparameterization as in step 6, we choose to compute it once at the beginning of the algorithm for the sake of increased computational efficiency.  

In this paper, we provide a validation of GORA-S using skeleton sequences from the NTU RGB+D Action Recognition Dataset \cite{shahroudy2016ntu}. The following sections describe the specific formulation of GORA-S used in our experiments.

\begin{figure*}[!t]
\centering
\includegraphics[width = \textwidth]{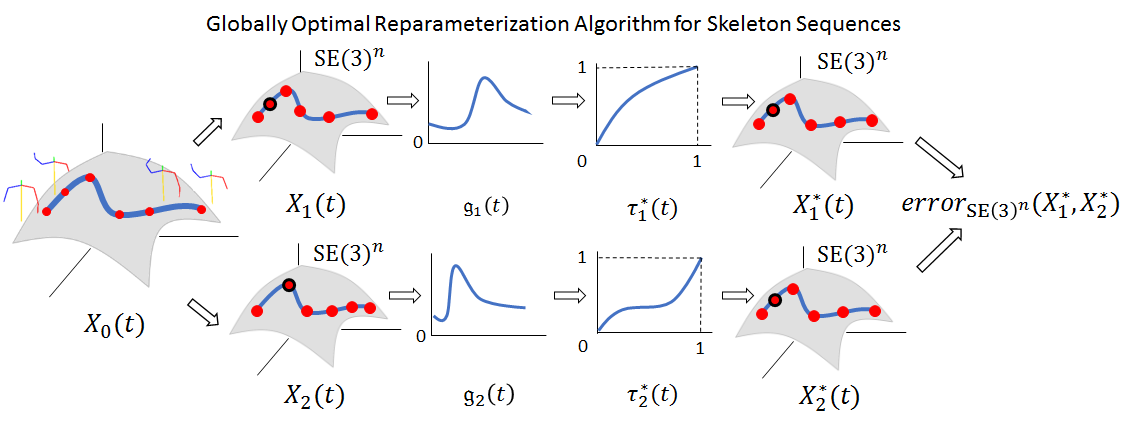}
\caption{The experimental work flow for evaluating the computational efficiency and accuracy of GORA-S in matching skeleton sequences: A template skeleton $X_0(t)$ is selected and parameterized with respect to two randomly generated functions in the TRG (e.g. $\tau_1(t)$ and $\tau_2(t)$) to create two skeletons $X_1(t), X_2(t) \in \skls$, identical up to their temporal alignment. For each input signal, $\tau^*(t)$ is computed from $\mathfrak{g}(t)$ and used to reparameterize the skeleton as in Algorithm \ref{algo1}. The algorithm then returns the error between the UST reparameterizations computed using the metric in \eqref{d_skl}.}
\label{fig:demo_skl_reparam}
\end{figure*}

\subsection{Formulation of $\mathfrak{g}(t)$}
For skeleton sequences, a natural choice for the definition of $\mathfrak{g}(t)$ consistent with~(\ref{cost}) is based on the body velocities of the joint trajectories in $\SE(3)$ comprising the signal. For a skeleton with $n$ joint trajectories as in~(\ref{skelly}), we defined $\mathfrak{g}(t)$ as 
\begin{equation}
\mathfrak{g}(t) = \mathfrak{g}(t; \dot{X}(t)) = \sum_{j = 1}^{n}\norm{g_{j}^{-1}\frac{\partial g_{j}}{\partial \tau} }_W^{2}. \label{g_tau}
\end{equation}
where $\norm{\cdot}_W$ denotes the weighted Frobenius norm defined such that for any $A \in \mathbb{R}^{4 \by 4}$, \begin{align}
\norm{A}_W = \sqrt{\textrm{tr}\left(A^T W A\right)} \label{wnorm} \end{align} given a symmetric $4 \times 4$ matrix $W$. Here, we defined $W$ as \begin{align} W = \begin{bmatrix} J & {\bf 0} \\ {\bf 0}^T & m \end{bmatrix} \label{Wmat}\end{align} with $m =  1$ and \begin{align} J = \frac{1}{2}\textrm{tr}(I) \mathbb{I} - I \label{thatJ} \end{align} where $I$ is the $3 \times 3$ diagonal inertia tensor corresponding to a solid sphere of unit mass and $\mathbb{I}$ denotes the $3 \times 3$ identity matrix.  Further details of this norm can be found in \cite{chirikjian2016harmonic}.

\subsection{Interpolation on $\skls$} \label{interp}
Given a set of $T$ time instances $\{t_i\}$, the corresponding values of a skeleton sequence $$\{X(t_i)\} = \{(g_1(t_i), \ \ldots, g_n(t_i))\}$$ and the values of its temporal derivative $$\{\dot{X}(t_i)\} = \{(\dot{g}_1(t_i), \ \ldots, \dot{g}_n(t_i))\},$$ we can construct a piecewise interpolating curve for each joint trajectory, $g_j(t), \ 1 \leq j \leq n$, such that $$\bar{X}(t) = (g_1(t), \ \ldots, g_n(t))$$ passes through $X(t_i)$ at time instance $t_i, 1 \leq i \leq T$, as follows \cite{Belta2002AnSP}: 

For a given joint trajectory $g_j \in \SE(3)$, we can define a cubic minimum acceleration curve in $\textrm{Aff}^+(4, \mathbb{R})$ by \begin{align*}
M(t) & = \begin{bmatrix} M_{3 \by 3}(t) & {\bf m}(t) \\ {\bf 0}^T & 1 \end{bmatrix} \\ & = M_3t^3 + M_2 t^2 + M_1 t + M_0, \quad t \in [t_i, t_{i+1}] \end{align*}
where \begin{align*}
& M_3 = 6 \ \frac{\dot{g}_j(t_i) + \dot{g}_j(t_{i+1})}{{(\Delta t)}^2} - 12\frac{\Delta x}{(\Delta t)^3} \\
& M_2 = \frac{\Delta v}{\Delta t} - M_3 \frac{t_i + t_{i+1}}{2} \\
& M_1 = \dot{g}_j(t_i) - M_3\frac{t_i^2}{2} - M_2 t_i \\
& M_0 = g_j(t_i) - M_3 \frac{t_i^3}{6} - M_2 \frac{t_i^2}{2} - M_1 t_i
\end{align*} and 
\begin{align*}
& \Delta t = t_{i+1} - t_i \\
& \Delta x = g_j(t_{i+1}) - g_j(t_{i}) \\
& \Delta v = \dot{g}_j(t_{i+1}) - \dot{g}_j(t_{i})
\end{align*}
Taking $J$ as defined in~(\ref{thatJ}) we can find the singular value decomposition of $M_{3 \by 3} (t) J$ to recover the matrices $ U(t), \Sigma(t), V(t)$ such that $$ M_{3 \by 3}(t) J = U(t) \Sigma (t) V^{H}(t).$$ Then, the curve interpolating the joint trajectory $g_j \in \SE(3)$ on the interval $[t_i, t_{i+1}]$ is given by
\begin{align*} g_j(t) = \begin{bmatrix} R(t) & {\bf r}(t) \\ {\bf 0}^T & 1 \end{bmatrix}, \quad t \in [t_i, t_{i+1}] \end{align*} where
\begin{align*} & R(t) = U(t) V^{H}(t) \in \SO(3) \\ & {\bf r}(t) = {\bf m}(t) \in \mathbb{R}^3 \end{align*}
				
\subsection{Numerical differentiation of skeleton sequences}
We use Fornberg's method to numerically calculate $\dot{X}(t)$ for skeleton signals from the NTU Action Recognition dataset \cite{fornberg_generation_1988,fornberg_classroom_1998}. This method can compute derivatives on both regularly and irregularly spaced grids, making it potentially advantageous in real world scenarios in the case where an RGB+D camera might fail to track a skeleton at certain time instances. In our experiments, we use Bjorn Dahlgren's \verb| finitediff | package in Python 2.7, which enables efficient numerical differentiation over arrays via Fornberg's method \cite{bjorn_dahlgren_2018_1299173}. 

\subsection{Error metric}
Given two skeleton sequences, \begin{gather*} X_1(t) = (g_1(t), \, \ldots, g_n(t)) \\ 
X_2(t) = (h_1(t), \ \ldots, h_n(t)), \end{gather*} both in $\skls$, we defined the element-wise distance between them at an arbitrary time instance $t_0$ as \begin{gather}
d_{\skls}(X_1(t_0), X_2(t_0)) = \nonumber \\ \frac{1}{n}\sum_{j=1}^{n} \norm{\log_{\SE(3)} \left(\left[g_j(t_0)\right]^{-1} \left[h_j(t_0)\right]\right)}_{W}, \label{d_SE_n}\end{gather} where $\log_{\SE(3)}(\cdot)$ denotes the logarithm operation which maps elements in $\SE(3)$ to their corresponding elements in the Lie algebra, $\se(3)$ \cite{Murray:1994:MIR:561828}.  Similarly, we defined the average distance or error between two skeletons across all time instances $\{t_i\}$ to be \begin{align}
error_{\skls} = \frac{1}{T}\sum_{i = 1}^{T} d_{\skls}(X_1(t_i), X_2(t_i)), \label{d_skl}
\end{align}
where $T$ is the total number of the time instances.

\section{Algorithm Performance and Comparisons}
\label{Comparisons}
This section summarizes our comparisons between GORA-S and that of the DTW and FastDTW \cite{salvador2007toward} algorithm in the context of action recognition with skeleton sequences. Specifically, we evaluated the performance of each of the above algorithms in terms of both accuracy in matching skeleton sequences representing the same humanoid action and computational efficency. All comparisons are performed in Python 2.7 and the DTW and FastDTW implementations used in our experiments were from the official Python package \cite{tanida_fastdtw}. The experiments were performed on an Intel Core i7-7600U CPU @ 2.80GHz. 

\subsection{Pre-processing}
Due to the large size of the NTU Action Recognition dataset, we randomly chose 50 skeleton sequences from each of the drinking water (A001), clapping (A010), cheering (A022), and waving (A023) action classes to create a pool of skeletons with which to compare algorithms. It is important to note that for the sake of sampling consistency, we trimmed all skeleton sequences in this pool such that each sequence depicted only a \textit{single instance of an action being performed}. For example, skeleton sequences of a person waving multiple times were trimmed to show only a single wave. Since each of the four chosen action classes were characterized by upper body movements, we pruned each skeleton in the pool to include only the joints comprising the torso, arms, and head to eliminate any added noise from irrelevant joints. We also excluded any joints that did not contain rotational components in \SO(3). Additionally, we applied a normalization procedure to each skeleton to account for differences in physical size between subjects and body orientation with respect to the camera coordinates. The complete details of the procedure can be found in \cite{shahroudy2016ntu}.

\subsection{Comparison regime}

\begin{figure}[!t] 
\centering
\begin{subfigure}{\columnwidth}
\includegraphics[width=\textwidth]{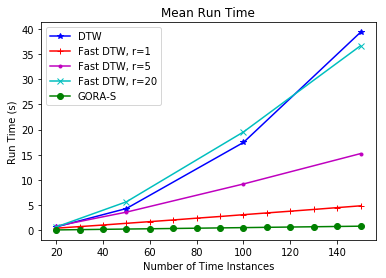}
\caption{Mean run time}
\label{fig:skl_run_time}
\end{subfigure}
\vskip \baselineskip
\begin{subfigure}{\columnwidth}
\includegraphics[width=\textwidth]{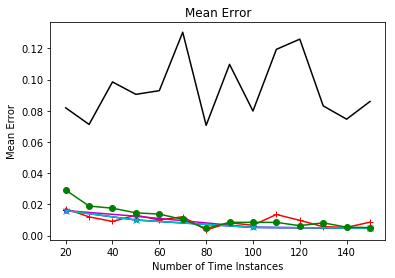}
\caption{Mean error}
\label{fig:skl_error}
\end{subfigure}
\vskip \baselineskip
\begin{subfigure}{\columnwidth}
\includegraphics[width= \textwidth]{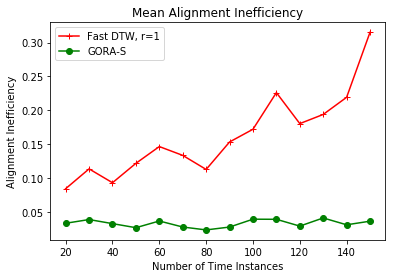}
\caption{Mean alignment inefficiency: GORA-S vs. FastDTW}
\label{mean_AI}
\end{subfigure}
\caption{Algorithm performance: skeleton sequences.} \label{skl_comp}
\end{figure}

We compared the performance of GORA-S with the DTW algorithm and implementations of the FastDTW algorithm with radii of 1, 5, and 20. The procedures with which we performed action recognition comparisons with skeleton sequences are as follows: For a given number of time instances, we randomly selected 50 different template skeletons from the pool. For each template skeleton, two initial parameterizations in the TRG were randomly generated and used to parameterize the original signal, creating 50 pairs of input signals with different temporal alignments, which were then fed to GORA-S and the DTW and FastDTW algorithms.

To ensure fair comparisons between algorithms, we use a modified version of GORA-S designed for pairwise comparison of two skeleton sequences, outlined in Fig. \ref{fig:demo_skl_reparam}. This version accepted two input skeletons, $X_1(t), X_2(t) \in \skls$, computed in parallel their respective UST reparameterizations, i.e. $X_1^*(t)$ and $X_2^*(t)$, as in Algorithm \ref{algo1} and outputted the average error between the two UST reparameterizations given by~(\ref{d_skl}). An example of this approach can be seen in the supplementary video accompanying this paper. Similarly, we implemented the DTW and FastDTW algorithms such that the metric given in~(\ref{d_SE_n}) was used to compute the element-wise distance between skeletons at arbitrary time instances. Furthermore, we normalized the accumulated cost error output by the DTW and FastDTW algorithms under~(\ref{d_SE_n}) by dividing it by the length of the optimal warping path. Run time comparisons were performed using the \verb|clock| module in Python's \verb|time| package. Given two input signals, we defined the run time (what we called computational efficiency) to be the time it took each algorithm to output the error between two skeleton sequences.  

To quantify the relationship between run time and accuracy, we introduce a quantity we call the \textit{alignment inefficiency}, denoted as $\cal{I}$. For an arbitrary signal alignment algorithm and pair of input skeletons, letting $E_0$ denote the initial error between the skeletons given by~(\ref{d_skl}), $E_f$ denote the error between skeletons found by the algorithm, and $T_R$ denote the run time of the algorithm, we can define the alignment inefficiency as \begin{align}
{\cal{I}} = \frac{E_f T_R}{E_0} \label{AI}. \end{align} 

The alignment inefficiency gives a spatially normalized measure of an algorithm's performance that equally weights run time and accuracy. For a given pair of skeleton sequences, an algorithm with a small run time and a larger error would have a similar alignment inefficiency as that of another algorithm with a larger run time and a smaller error. However, a `better' algorithm with both a small run time and small error would have the smallest alignment inefficiency. In addition, the rate of change of the alignment inefficiency with respect to the temporal sampling frequency of the input skeleton sequences provides a measure of the sensitivity of the algorithm's performance to the degree of coarseness or fineness in the sampling of the signals.

\subsection{Results}

Fig. \ref{skl_comp}. compares the performance of GORA-S and the DTW and FastDTW algorithms. Fig. \ref{fig:skl_run_time}. shows the mean run time of each algorithm from 20 to 150 time instances. For skeleton sequences, as the total number of time instances increases,  DTW's run time grows quadratically (i.e. $O(T^2)$ complexity) while all iterations of the FastDTW algorithm and GORA achieve linear complexity (i.e. $O(T)$). However GORA-S's run time is less than that of all DTW methods, and its complexity grows more slowly than the fastest implementation of FastDTW (radius $=$ 1). 

Fig. \ref{fig:skl_error}. shows the mean error between skeleton pairs identical up to their temporal alignments given by each algorithm from 20 to 150 time instances. The accuracy of GORA-S, in the sense that the computed error skeleton pairs representing the same action is small, is comparable to that of the the DTW algorithm and all implementations of the FastDTW algorithm, though slightly poorer when comparing very coarsely sampled skeleton sequences. It was often the case that the DTW algorithm and the implementations of the FastDTW algorithm gave identical errors, since it is possible for the FastDTW algorithm to construct the same accumulated cost matrix as the DTW algorithm. 

Fig. \ref{mean_AI}. compares the mean alignment inefficiencies of GORA-S and the FastDTW algorithm with radius $=$ 1 across skeleton pairs with 20 to 150 time instances. In the opinion of the authors, the disparity between the mean alignment inefficiency of the FastDTW algorithm and that of GORA-S is especially significant. While GORA-S is slightly less accurate than FastDTW with radius $=1$ when comparing coarsely sampled skeleton sequences, its lower alignment inefficiency implies that the loss in accuracy is outweighed by the relative increase in computational efficiency gained due to its faster run time.

	More importantly, GORA-S's alignment inefficiency remains approximately constant as the sampling frequency of the input skeletons increases. This implies that the overall performance of GORA-S is approximately invariant with respect to the sampling frequency of the skeleton sequences. In other words, any increases in run time when comparing more finely sampled skeletons are offset by directly proportional decreases in the computed error and vice-versa. Moreover, this suggests that GORA-S has the potential to be an extremely `versatile' algorithm that can perform efficiently in a variety of roles. Its faster run times and competitive accuracy relative to DTW methods when comparing coarsely sampled skeletons make it well-suited for real-time action recognition while its ability to remain efficient when comparing finely sampled skeletons could make it an effective algorithm for deep learning applications.  However, it is important to keep in mind that these results represent only an initial validation of GORA-S using skeletons depicting four elementary humanoid actions and that further analysis is needed to properly contextualize the strengths and weakness of the algorithm. 

\section{DISCUSSION}
As noted in \cite{mitchelruan2018}, an important difference between GORA-S and the DTW and FastDTW algorithms is the reliance of GORA-S on interpolation to recover the UST reparameterization of the input skeleton. In the context of action recognition with skeleton sequences, this gives GORA-S an advantage over DTW methods since the Frobenius norm~(\ref{wnorm}) is expensive to compute relative to its vector analog. By interpolating, GORA-S only has to compute this norm $3nT$ times ($nT$ times in both calculations of~(\ref{g_tau}) and $nT$ times in~(\ref{d_skl}), where $n$ is the number of joint trajectories comprising each skeleton and $T$ is the total number of time instances. On the other hand, all DTW methods have to compute this error between $n \cdot O(T)$ and $n \cdot O(T^2)$ times.

This advantage is substantial enough that we can interpolate inside GORA-S to recover UST reparameterizations using the method described in Section \ref{interp}, which requires computing the singular value decomposition factorization of each of the $3 \times 3$ matrices in $\textrm{GL}(3, \mathbb{R})$ at the desired points of evaluation along the minimum acceleration cubics in $\textrm{Aff}^+(4, \mathbb{R})$ between temporally adjacent joints $2nT$ times, while remaining faster than all implementations of the FastDTW algorithm. A simpler and less computationally expensive method could be used to interpolate the joint trajectories of skeletons in $\skls$, such as the well known minimum geodesic method on $\SE(3)$ \cite{zefran1996choice}, however the increase in computational efficiency would likely be offset by a significant decrease in accuracy.

	We have not yet examined the performance of GORA-S relative to DTW methods when external noise is added to skeleton sequences. As a next step, we plan to explore GORA-S's capabilities in providing a foundation for a more robust action recognition algorithm able to minimize or eliminate noise and nuisance parameters in skeleton sequences while simultaneously reparameterizing them to a UST \cite{chirikjian2017signal}. The development of such an algorithm able to inherently compensate for perturbations such as noise or motion artifacts while maintaining a linear complexity similar to that of GORA-S would mark an important milestone toward the goal of robust robotic action recognition of human motions in real-time.


\section{CONCLUSIONS}
In this paper, we introduced a variant of the \textit{Globally Optimal Reparameterization Algorithm} for signal alignment and action recognition with skeleton sequences, which we call GORA-S.  This algorithm reparameterizes skeletons to a \textit{universal standard timescale} (UST), allowing for element-wise comparisons between  skeletons at each time instance with a linear time complexity of $O(T)$. Additionally, we review the parameters and numerical techniques used in its application.

Our experimental results suggest that GORA-S has the potential to become a viable alternative to DTW methods for signal alignment and action recognition with skeleton sequences. In particular, we show that the computational complexity of GORA-S is less than that of the FastDTW algorithm with radius $=$ 1 and that it attains a competitive degree of accuracy in matching skeleton sequences. More importantly, GORA-S displays a favorable balance between speed and accuracy that remains invariant under changes in the temporal sampling frequency of the input skeletons, suggesting it has the potential to be a versatile algorithm well-suited for a variety of different action recognition related tasks.

\section{ACKNOWLEDGMENTS}
The authors would like to thank Dr. Jin Seob Kim for useful discussions, Mr. Can Kocabalkanli and Ms. Yuqing (Eva) Pan for pictures of the NAO robot, and Mr. Mark Shifman for help in pre-processing data. This work was performed under National Science Foundation grant IIS-1619050 and Office of Naval Research Award N00014-17-1-2142.

\bibliography{goras_ref}
\bibliographystyle{IEEEtran}

\end{document}